\documentclass[10pt,twocolumn,letterpaper]{article}

\usepackage{cvpr}
\usepackage{times}
\usepackage{epsfig}
\usepackage{graphicx}
\usepackage{amsmath}
\usepackage{amssymb}

\usepackage{subfigure}
\usepackage{booktabs} 
\usepackage{diagbox}
\usepackage{multirow}
\usepackage{ifthen}
\usepackage{lineno}
\usepackage{ifthen}
\usepackage{cases}
\usepackage{footnote}
\usepackage{lipsum}
\makesavenoteenv{table}

\usepackage{kotex}
\usepackage{color}
\DeclareMathAlphabet{\pazocal}{OMS}{zplm}{m}{n}

\definecolor{greyblue}{rgb}{0.1,0.6,0.5}

\definecolor{purple}{rgb}{0.626,0.125,0.941}

\usepackage[pagebackref=true,breaklinks=true,letterpaper=true,colorlinks,bookmarks=false]{hyperref}

\cvprfinalcopy 

\def\cvprPaperID{5963} 

\ifcvprfinal\fi
\begin{document}

\title{Deep Relational Reasoning Graph Network for Arbitrary Shape Text Detection}


\author{Shi-Xue Zhang$ ^1 $, Xiaobin Zhu$ ^1 $, Jie-Bo Hou$ ^1 $, Chang Liu$ ^1 $ \\ Chun Yang$ ^1 $, Hongfa Wang$ ^4 $, Xu-Cheng Yin$ ^{1,2,3} $\thanks{Corresponding author.}\\
	{\tt\small $ ^1 $School of Computer and Communication Engineering, University of Science and Technology Beijing}\\
	{\tt\small $ ^2 $Institute of Artificial Intelligence, University of Science and Technology Beijing}\\
	{\tt\small $ ^3 $USTB-EEasyTech Joint Lab of Artificial Intelligence, \tt\small $ ^4 $Tencent Technology (Shenzhen) Co. Ltd}\\
	{\tt\small zhangshixue111@163.com, \{zhuxiaobin, chunyang, xuchengyin\}@ustb.edu.cn}\\ {\tt\small houjiebo@gmail.com, lasercat@gmx.us,  hongfawang@tencent.com}
}

\maketitle
\thispagestyle{empty}
\begin{abstract}
	Arbitrary shape text detection is a challenging task due to the high variety and complexity of scenes texts.
	In this paper, we propose a novel unified relational reasoning graph network for arbitrary shape text detection. In our method, an innovative local graph bridges a text proposal model via Convolutional Neural Network (CNN) and a deep relational reasoning network via Graph Convolutional Network (GCN), making our network end-to-end trainable. To be concrete, every text instance will be divided into a series of small rectangular components, and the geometry attributes (\eg, height, width, and orientation) of the small components will be estimated by our text proposal model. Given the geometry attributes, the local graph construction model can roughly establish linkages between different text components.
	For further reasoning and deducing the likelihood of linkages between the component and its neighbors, we adopt a graph-based network to perform deep relational reasoning on local graphs. Experiments on public available datasets demonstrate the state-of-the-art performance of our method. Code is available at \url{https://github.com/GXYM/DRRG}.
\end{abstract}

\section{Introduction}
\begin{figure}[ht]
	\begin{minipage}[t]{0.98\linewidth}
	\includegraphics[width=1\linewidth]{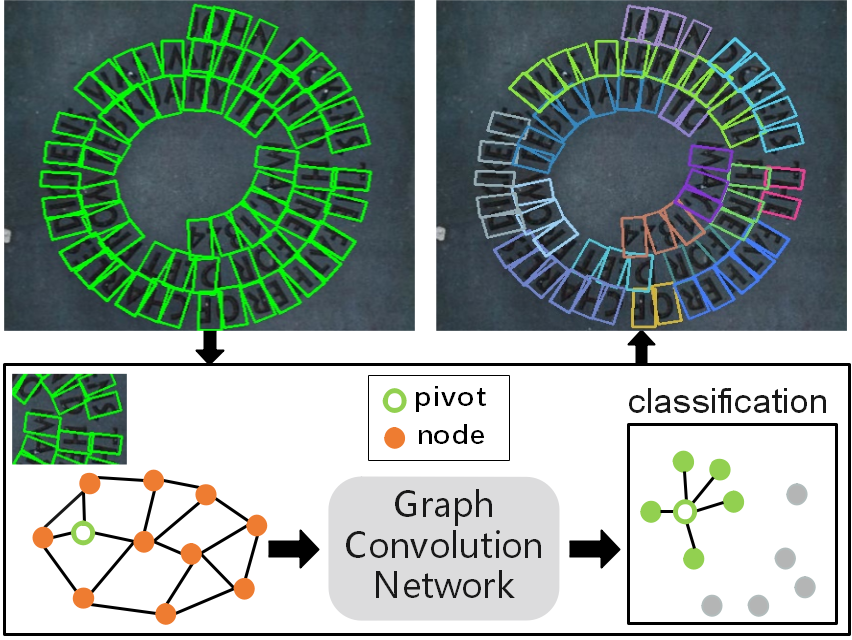}
	\caption{Illustration of relational reasoning:
		 Generating local graphs based on the geometry attributes of text components;
		  inferring linkage likelihood via GCN; finally grouping node classification results into text.
	}
	\label{fig:fig1}
	\end{minipage}%
	\vspace{-0.5em}	
\end{figure}

Scene text detection has been widely applied in various applications, such as online education, product search, instant translation, and video scene parsing \cite{Yin-Z, Yin-V}.
With the prosperity of deep learning, text detection algorithms \cite{CTPN, EAST,SegLink,RRPN} have achieved impressive performance in controlled environments where text instances have regular shapes or aspect ratios. However, because of the limited text representation forms, pioneer works tend to fail in detecting texts with arbitrary shapes.  Recently, some methods, \eg, TextSnake \cite{TextSnake} and CRAFT \cite{CRAFT}, try to solve this problem with the Connected Component (CC) strategy. However, these methods haven't fully explored the abundant relations between text components, which can benefit the aggregation of text components for final text instance.

In the CC-based method, one essential task is to excavate the rational relations between separated character/component regions for linking them into holistic text instances. The existing methods usually use pre-defined rule, link map or embedding map to group the detected components into text instance. Generally speaking, grouping the text components with learned link relationship or embedding relationship is more robust than using pre-defined rules, especially in the cases of long and curve texts. From our key observations and experiments, deep relational reasoning for mining the stable relations between these component regions can greatly enhance the performance of arbitrary shape text detection. The link or embedding based methods \cite{SegLink, CVPR19_LSA} usually uses CNNs to deduce the linkage of separate components, but the separated components are actually non-Euclidean data and CNNs are not good at processing non-Euclidean data.
Therefore, the simple link map or embedding map is inadequate for learning stable relations between two non-adjacent components. The non-Euclidean data can be represented with graph, so we can transform the separate text components into graphs. As shown in Fig.$ \, $\ref{fig:fig1}, we regard one text component as a node. Hence, we can select a node as a pivot and connect it with surrounding nodes into a local graph, as described in Sec.$ \, $\ref{section:local_graph}. The context information contained in local graphs (edges among the nodes) is informative for estimating the linkage likelihood between pivot and other nodes. It’s a consensus that graph network has innate advantage for deducing relationships between nodes on the graph. Recently, the GCN based methods have achieved remarkable performance in clustering face \cite{CVPR19_Linkage} and global reasoning for various tasks \cite{CVPR19_GloRe}. Highly motivated by the works in \cite{CVPR19_Linkage, CVPR19_GloRe}, we apply a graph convolution network to perform deep reasoning on local graphs to deduce deep linkage likelihood between components and corresponding neighbors for arbitrary shape text detection.

In this paper, we propose a novel unified deep relational reasoning graph network for arbitrary shape text detection. According to CTPN \cite{CTPN} and TextSnake \cite{TextSnake}, we divide every text instance into text components, and propose a text proposal network to estimate the geometry attributes of these text components. To group the generated components, we adopt a graph-based network to perform deep relational reasoning and inferring the linkage relationship using the geometry attributes of components and neighbors. In addition, a local graph is designed to bridge the text proposal network and relational reasoning network, making our network end-to-end trainable. Finally, we group the detected text components into holistic text instances according to the relational results. 

In summary, the main contributions of this paper are three-fold:
\begin{itemize}
\item We propose a novel unified end-to-end trainable framework for arbitrary shape text detection, in which a novel local graph bridges a CNN based text proposal network and a GCN based relational reasoning network.
\item To the best of our knowledge, our work presents one of the very first attempts to perform deep relational reasoning via graph convolutional network for arbitrary shape text detection.
\item The proposed method achieves the state-of-the-art performance both on polygon datasets and quadrilateral datasets.
\end{itemize}

\section{Related Work} \label{Related_Work}


\textbf{Regression-Based Methods.} Methods of this type rely on a box-regression based object detection frameworks with word-level and line-level prior knowledge ~\cite{RRPN,textboxes++,RRD,EAST}. Different with generic objects, texts
are often presented in irregular shapes with various aspect ratios. To deal with this problem, RRD ~\cite{RRD} adjusts anchor ratios of SSD ~\cite{SSD} for accommodating the aspect ratio variations in irregular shapes. Textboxes++ ~\cite{textboxes++}
modifies convolutional kernels and anchor boxes to effectively capture various text shapes. EAST ~\cite{EAST} directly inferences pixel-level quadrangles of word candidates without anchor mechanism and proposal detection. Although regression-based methods have achieved good performance in quadrilateral text detection, they often can't well adapt to arbitrary shape text detection.


\textbf{Segmentation-Based Methods.}
Methods of this type ~\cite{PixelLink, CVPR19_PSENet, CVPR19_LSA,TextField, TextSnake} mainly draw inspiration from semantic segmentation methods and detect texts by estimating word bounding areas.
In PixelLink$ \, $~\cite{PixelLink}, linkage relationships between a pixel and its neighboring pixels are predicted for grouping pixels belonging to same instance. To effectively distinguish adjacent text instances, PSENet ~\cite{CVPR19_PSENet} adopts a progressive scale algorithm to gradually expand the pre-defined kernels. Tian \etal. ~\cite{CVPR19_LSA} considered each text instance as a cluster and perform pixel clustering through an embedding map. TextField ~\cite{TextField} adopts a deep direction field to link neighbor pixels and generate candidate text parts. However, the performances of these methods are strongly affected by the quality of segmentation accuracy.
\begin{figure*}[htbp]
	\vspace{-0.0em}
	\begin{center}
	\includegraphics[width=0.95\linewidth]{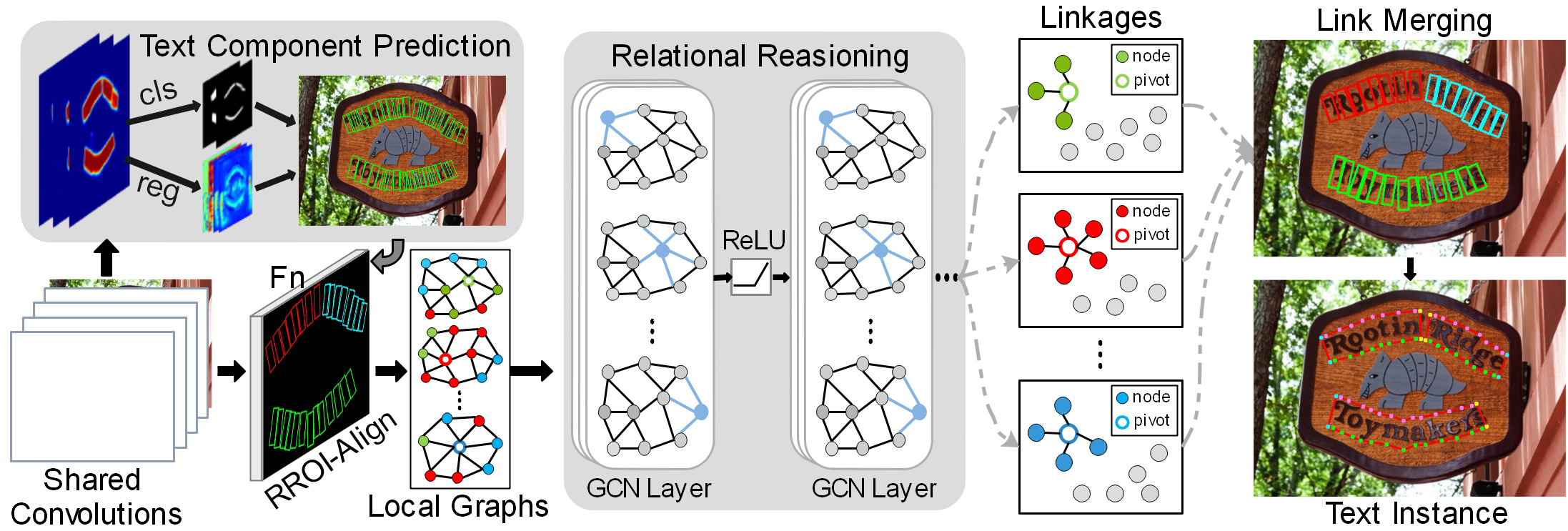}
	\caption{Overview of our overall architecture. Our network mainly consists of five components, \ie, shared convolutions, text component prediction, local graphs, relational reasioning, and Link Merging.}
	\label{fig:fig2}
	\end{center}
	\vspace{-1.8em}
\end{figure*}

\textbf{CC-Based Methods.}	The CC-based methods usually detect individual text parts or characters firstly, followed by a link or group post-processing procedure for generating final texts. CC-based methods $ \ $~\cite{MSER14, MSER15, SWT15, Yin-M} had been widely used in traditional scene text detection methods before the popularity of deep learning.
In the era of deep learning, CC-based methods have also been
extensively studied ~\cite{CTPN, SegLink,SegLink++, CRAFT, TextDragon}. CTPN ~\cite{CTPN} uses a modified framework of  Faster R-CNN ~\cite{Faster-rcnn} to extract horizontal text components with a fixed-size width for easily connecting dense text components and generating horizontal text lines. SegLink ~\cite{SegLink} decomposes every scene text into two detectable elements, namely segment and link, where the link indicates that a pair of adjacent segments belong to the same word. CRAFT ~\cite{CRAFT} detects the text area by exploring each character and affinity between characters. TextDragon ~\cite{TextDragon} first detects the local area of the text, and then groups these bounding boxes according to their geometric relations.


\textbf{Relational Reasoning.} CC-based  methods are usually robust for long or non-quadrilateral text, but the performance of these methods are strongly depends on the robustness of grouping or linkage results. Text pixels are clustered by learning the linkage relationship between a pixel and its neighboring pixels in ~\cite{PixelLink} . In ~\cite{CVPR19_LSA}, embedding features are used to provide instance information and to generate the text area. CRAFT ~\cite{CRAFT} predicts character region maps and affinity maps by weakly-supervised learning. The region map is used to localize characters, and the affinity map is used to group characters into a single instance. These methods are based on the CNNs, which cannot directly capture relations between distant component regions for the limitation of local convolutional operators.
Recently, Wang \etal. ~\cite{CVPR19_Linkage} proposed a spectral-based GCN to solve the problem of clustering faces, where the designed GCN can rationally link different face instances belonging to the same person in complex situations.

\section{Proposed Method} \label{Proposed_Method}

\subsection{Overview}
\begin{figure}[ht]
	\vspace{-1.0em}	\begin{center}
	\includegraphics[width=0.95\linewidth]{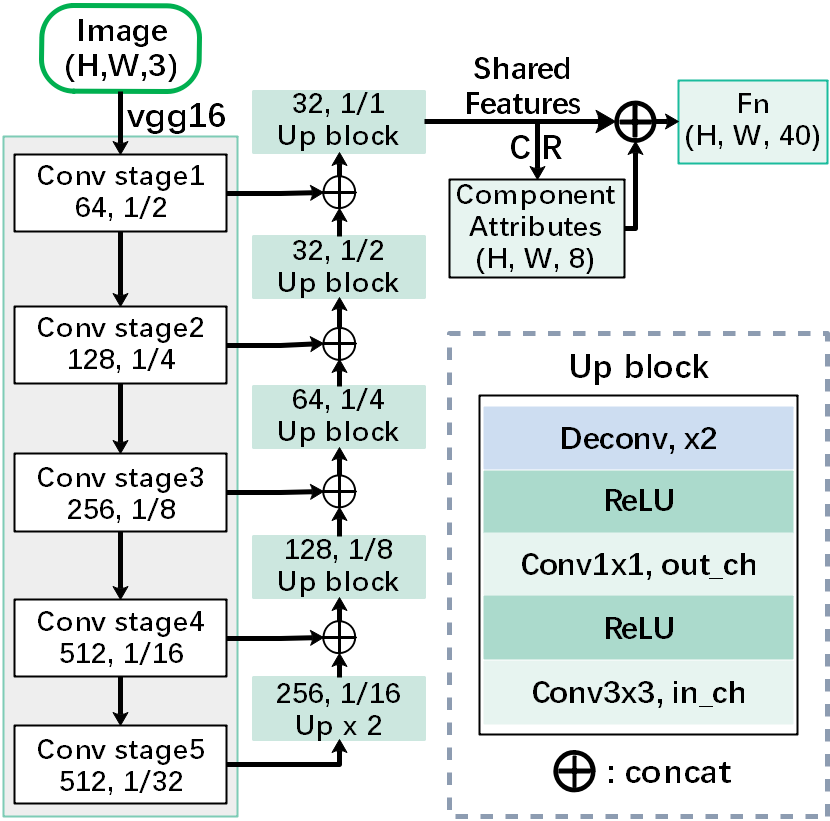}
		\caption{Architecture of shared convolutions, where $ CR $ represents classification and regression operation in text component prediction, and details are listed in Eq.$ \ $\ref{CR}.
		}
	\label{fig:fig3}
	\end{center}%
	\vspace{-2em}
\end{figure}

The framework of our method is illustrated in Fig.$\,$\ref{fig:fig2}.
The text component proposal network and the deep relational reasoning graph network share convolutional features, and the shared convolutions use the VGG-16 \cite{vgg} with FPN \cite{FPN} as backbone, as shown in Fig.$\,$\ref{fig:fig3}. The text proposal network uses the shared features to estimate geometric attributes of text components. After obtaining the geometry attributes, the local graph can roughly establish linkages between different text components. Based on local graphs, the relational reasoning network will further infer the deep likelihood of linkages between the component and its neighbors. Finally, text components will be aggregated into holistic text instance according to the reasoning results.

\subsection{Text Component Prediction}\label{TextPredict}
\begin{figure}[tbp]
	\vspace{-0.0em}
	\begin{center}
		\includegraphics[width=0.98\linewidth]{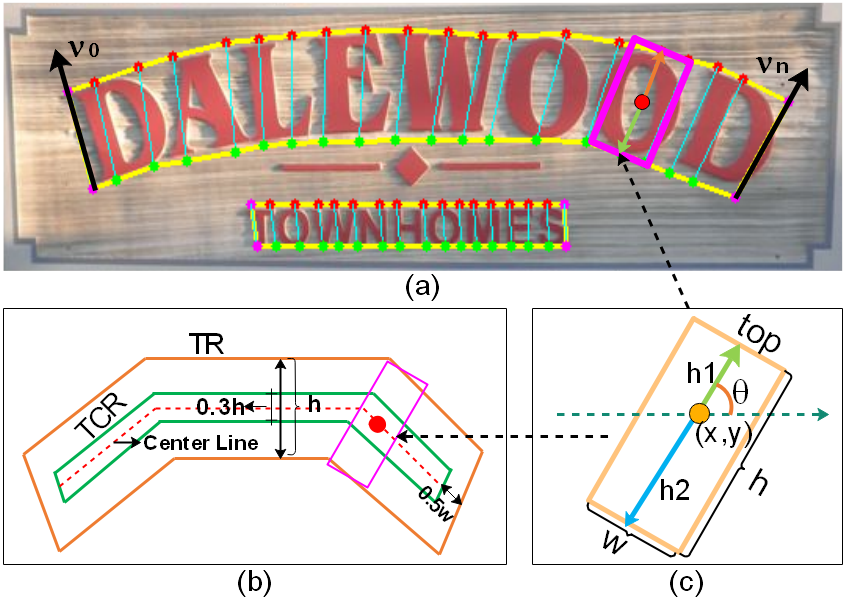}
		\caption{Illustration of the proposal of text component: (a) Generating text component; (b) Extracting text center region; (c) Calculating geometry attributes.
		}
	\label{fig:fig4}
	\end{center}%
	\vspace{-2.0em}
\end{figure}

In our work, each text instance is constructed by a series of ordered rectangular components, as shown in Fig.$ \, $\ref{fig:fig4} (a). 
And each text component $ D $ is associated with a group of geometry attributes, \ie,
$ D= (x , y, h , w , \cos\theta, \sin\theta) $, in which $ x$ and $y $ are the axis of text box; $ h$ and $ w  $ are the height and the width of the component; $ \cos \theta $ and  $ \sin \theta $ indicate the orientation of the text component. The $ h $ is the sum of $ h1 $ and $ h2 $, as shown in Fig.$ \,$\ref{fig:fig4} (c). The $ w $ is obtained by a linear transformation on $ h $, which is computed as
\begin{equation}\small\label{h_compute}
w_i =
\begin{cases}
w_{min},\; h_i<=2 \cdot w_{min},\\
h_i/2, \quad 2 \cdot w_{min}<h_i<2 \cdot w_{max},\\
w_{max},\; h_i>=2 \cdot w_{max},
\end{cases}
\end{equation}
where $ h_i $ denotes the height of the $ i $-th text component. In experiments, we empirically set $ w_{min}=8 $  and $ w_{max}=24 $.

In order to define the orientation of text components and extract the text center region (TCR) easily, we use the method in \cite{TextSnake} to calculate the head and tail of text region, as shown with the black arrow in Fig.$ \, $\ref{fig:fig4} (a). Text region is divided into a series of ordered quadrilateral regions along the long side (indicated by yellow lines as shown in Fig.$ \, $\ref{fig:fig4} (a)), so we can obtain two groups of points $ P1 =\{tp_0, tp_1 ,... , tp_i,..., tp_n \} $ and  $ P2=\{bp_0, bp_1 ,... , bp_i,..., bp_n \} $ . The line marked with red points is the top line and green points is the bottom. In our approach, we need to clearly define the top and bottom of each text instance, according to the following criterion:
\begin{equation}
p = \sum_{i=0}^{n} sin(v_i), v_i \in V,
\end{equation}
where $ V $ ($ V = \{tp_0-bp_0 ,... , tp_i-bp_i,..., tp_n-bp_n \} $) is a group of vertexes ($ tp_i $  is the center of the top line and $ bp_i $ is the center of the bottom line). If $ p>=0 $, $ P1 $ is top and $ P2 $ is bottom, else $ P1 $ is bottom and $ P2 $ is top. The angle of vector $ v_i $ indicates the orientation $ \theta  $ of text component.

TCR is obtained by shrinking text region (TR), as shown in Fig.$ \, $\ref{fig:fig4} (b). First, we compute the text center line,  Then, we shrink the two ends of center line by $ 0.5w $ end pixels, making it easy for the network to separate adjacent text instances and reduce the computation cost of NMS. Finally, we expand the center line area by $ 0.3h $. After extracting shared features, two convolution layers are applied to predict the attributes of the text component as
\begin{equation}\label{CR}
CR = conv_{1\times1}(conv_{3\times3}(F_{share})),
\end{equation}
where $ CR \in \Re^{h \times w \times 8} $, with 4 channels for the classification logits of TR/TCR, and $ 4 $ channels for
the regression logits of  $h_1, h_2, \cos\theta, $ and  $\sin\theta $. The final predictions are obtained by softmaxing TR/TCR and regularizing $ \cos\theta $ and  $ \sin\theta $ for squaring sum equals 1 \cite{TextSnake}. Final detection results are produced by threshold and locality-aware NMS on the positive samples.

\textbf{Detection Loss}. The text component prediction loss is consisted of two losses, and computed as
\begin{equation}
L_{det} = L_{cls} + L_{reg},
\end{equation}
where $  L_{reg} $ is a smooth $ L1 $ \cite{Faster-rcnn} regression loss and $  L_{cls} $ is a  cross-entropy classification loss. The classification loss is computed as
\begin{equation}
L_{cls} = L_{tr} + \lambda_1L_{tcrp} + \lambda_2L_{tcrn}\label{cls_tall},
\end{equation}
where $ L_{tr} $ represents the loss for TR; $ L_{tcrp} $ only calculates pixels inside TR and $ L_{tcrn} $ just calculates the pixels outside TR. $ L_{tcrn} $ is used to suppress noise of the background in TCR. In this way,  the obtained TCR can benefit post-processing steps. The OHEM \cite{OHEM} is adopted for TR loss, in which the ratio between the negatives and positives is set to 3:1. In our experiments, the weights $ \lambda_1 $ and
$ \lambda_2 $ are empirically set to $ 1.0 $ and $ 0.5 $, respectively.

Because the attributes of height and orientation are absent for non-TCR region, we only calculate regression loss for TCR region as followings:
\begin{gather}
L_{reg} = L_{h} + \beta (L_{sin}+ L_{cos}\label{reg_all}),\\
L_{sin} = smooth_{L1}(\hat{\sin\theta}-\sin\theta),\\
L_{cos} = smooth_{L1}(\hat{\cos\theta}-\cos\theta)\label{La},\\
L_{h} = \dfrac{1}{\Omega}\sum_{i \in \Omega}(\log{(h+1)}\sum_{k=0}^{2} smooth_{L1}(\dfrac{\hat{h_{ki}}}{h_{ki}} -1)),
\end{gather}
where $ h_{ki} $, $\sin \theta$ and $ \cos \theta $ are ground-truth values, and $\hat{h_{ki}} $, $\hat{\sin \theta}$ and $ \hat{\cos \theta} $ are the corresponding predicted values; the $ \Omega $ denotes the set of  positive elements in TCR; the $ h $ is the height of text component in ground truth. The weight $ \log{(h+1)} $ is beneficial for the height regression of large scale text component. The hyper-parameter $ \beta$ is set to $ 1.0 $ in our work.

\subsection{Local Graph Generation}\label{section:local_graph}
We estimate the linkage likelihood between two nodes (text components) based on their context information in a local graph. It is inefficient to construct a whole graph for each image because text component usually only has the possibility of connection with its neighbors. Therefore, we construct multiple local graphs for every image.
These local graphs generally contain a limited number of nodes, which will make it easy making relational reasoning high efficient.

We modify IPS \cite{CVPR19_Linkage}  to generate local graph, where pivot's neighbors up to $ h $-hop are used as nodes. In our work, we just use $ 2 $-hop as nodes for local graph. For clear explanation, $ V_p $ is used to represent the nodes in local graph $ G_p $ and $ p $ represents the pivot. The $ 1 $-hop neighbors of $ p $ consist of 8 nearest neighbors, and the $ 2 $-hop neighbors of $ p $ consist of 4 nearest neighbors. The high-order neighbors provide auxiliary information of the local structure of the context between a pivot and its neighbor \cite{CVPR19_Linkage}.  Here, we only consider the Euclidean similarity $ E_s $ between nodes for performing KNN operation, and $ E_s $ computed as
\begin{equation}
E_s =1 - D(p, v_i)/max(H_{m},W_{m}), v_i \in V_p,
\end{equation}
where $ D(p, v_i) $ is an L2 distance between $ p $ and $ v_i $, $ H_{m} $ is image height, and $ W_{m} $ is image width. To avoid gradient accumulation of easy samples caused by many identical graphs in training, the pivot $ p $ should satisfy the following criterion:

\begin{equation}
G_{iou} = \dfrac{G_p \cap G_q}{G_p \cup G_q}<\xi, \; p, q \in T,
\end{equation}
where $ G_p  $ and $ G_q $ are two local graphs; the pivot $ p$ and  $q $  are in the same text instance $ T $; $ G_p \cap G_q $  is the intersection of 1-hop neighbors of  $ G_p $ and  $ G_q $; $ G_p \cup G_q $  is the union of 1-hop neighbors of  $ G_p $ and $ G_q $. In our experiments, $ \xi $ is set to $ 0.75 $ . This strategy not only leads
to considerable acceleration, but also reduces the number of easy samples, yet keep the balance of hard and easy samples.

\subsection{Deep Relational Reasoning}
The text components in every image will be divided into multiple local graphs by local graph generation,
which consists of the pivot and its 2-hop neighbors. The rough linkage information contained in the local graph (edges among the nodes) is valuable for estimating the linkage likelihood between the pivot and its neighbors. For further reasoning and deducing the likelihood of linkage between the pivot and its neighbors, We adopt a specific graph-based neural network \cite{CVPR19_Linkage, GCN} to excavate the linkage relationships between the pivot and its neighbors based on local graph. The graph is usually expressed as $ g(X, A) $, and the graph convolutional network usually takes the feature
matrix $ X $ and the adjacency matrix $ A $ as the input of the network. Therefore, we need to extract the feature matrix $ X $ and compute matrix $ A $ for the local graph.

\begin{figure}[tbp]
	\vspace{-0.0em}
	\begin{center}
		\includegraphics[width=0.99\linewidth]{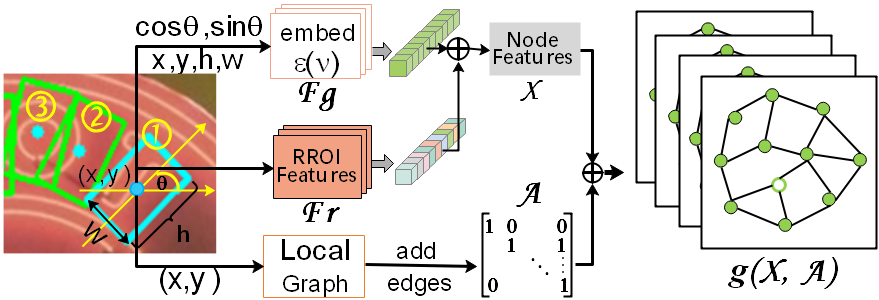}
		\vspace{0.8em}
		\caption{Illustration of $ g(X,A) $ generation. $ F_{r} $ represents the geometry features; $ X $  is node feature matrix; $ A $ is an adjacency matrix; $ g(X, A) $ represents the mathematical expression of local graph. The details of embedding operation in Eq.$ \, $\ref{embed1} and Eq.$ \, $\ref{embed2}. }
	\label{fig:fig5}
	\end{center}
	\vspace{-2.0em}
\end{figure}

\textbf{Node Feature Extraction.} The node features consist of two parts features, namely, RROI features and geometric features. In order to obtain the RROI features, we use the RRoI-Align layer which integrates the advantages of RoI-Align \cite{Mask_rcnn} and RRoI  \cite{RRPN} to extract the feature block of the input text component. To ensure the convergence ability of our model, we use the ground truth to generate the text component in training. Text components within the same text instance have similar geometric features. However, the RROI features will lose some geometric attributes, such as location information. Therefore, we should take these geometric attributes into consideration during node feature generation, as shown in Fig.$ \, $\ref{fig:fig5}. For one text component, we feed it with the feature maps $ F_{n} $ to RRoI-Align layer, and then a $ 1 \times 3 \times 4 \times C_r $ feature block is obtained, where $ F_{n} $ is illustrated in Fig.$ \ $\ref{fig:fig3}.  Afterwards, it will be reshaped to $ 1 \times 12 \cdot C_r $, namely $ F_r $. The geometric attributes of text component are embedded into high dimensional spaces according to the technique in \cite{Attention_Need, RegionFeatures}. The embedding is performed by applying $ sine $ and $ cosine $ functions of varying wavelengths to a scalar $ z $ as
\begin{gather}
\varepsilon_{2i}(z)=\cos(\dfrac{z}{1000^{{2i}/{C_\varepsilon}}}), i \in  (0,C_\varepsilon/2 - 1) \label{embed1}, \\ \varepsilon_{2i+1}(z)=\sin(\dfrac{z}{1000^{{2i}/{C_\varepsilon}}}), i \in  (0,C_\varepsilon/2 - 1), \label{embed2}
\end{gather}
where the dimension of the embedding vector $ \varepsilon(z) $ is $ C_\varepsilon $. As a result, each text component is embedded into a vector $ F_g $ with $ 6 \cdot {C_\varepsilon} $ dimension. Finally, $ F_r $ and  $ F_g $ will be concatenated together as node features.

\textbf{Node Feature Normalization.} We normalize the features of node by subtracting $ x_p $. It encodes the pivot $ p $ information into the features of a local graph and makes the relation reasoning network easily learn the linkage relationships between the pivot and its neighbors.
\begin{equation}
\bold F_p =[ ..., x_q-x_p, ...]^{T}, q \in V_p ,
\end{equation}
where $ x_p $ is the feature of the pivot $ p $;
the $ V_p $ denotes the node set on local graph and their features are $ \{x_q|q \in V_p\} $.

\textbf{Adjacency Matrix Generation.}
We use an adjacency matrix $ A_p \in \Re^{N*N} $ to represent the topological structure of local graph. For a node $ n_i \in V_p $ , we filter out the top $ u $ nearest neighbors $ U(n_i) $. For the node $ n_j \in U(n_i) $, we will set $ A_p(n_i,n_j) = 1 $. The hyper-parameter $ u $  is empirically set to 3 in our work.

\textbf{Graph Convolutions.} After obtaining the feature matrix $ X $ and the adjacency matrix  $ A $, we use a graph-based relational reasoning network to estimate the linkage relationships of the pivot and its neighbors based on the established graph. We modify the structure in \cite{CVPR19_Linkage, GCN}, and the graph convolution layer in our method can be formulated as
\begin{gather}
\bold Y^{(l)} = \sigma((\bold X^{(l)} \oplus \bold G\bold X^{(l)})\bold W^{l}),\\
\bold G = \bold{\tilde{D}}^{-1/2}\bold{\tilde{A}}\bold{\tilde{D}}^{-1/2},
\end{gather}
where $ \bold {X}^{(l)} \in \Re^{N \times d_{i}} , \bold { Y}^{(l)} \in \Re^{N \times d_{o}}  $, $ d_i/d_o $ is the dimension of input / output node features and $ N  $ is the number of nodes; $ \bold G $ is a symmetric normalized laplacian of
size $ N \times N $; the operator $ \oplus $ represents matrix concatenation; $ W^{(l)} $ is a layer-specific trainable weight matrix;
$ \sigma(\cdot) $ denotes a non-linear activation function; $ \tilde{A} = A + I_N $ is an adjacency matrix of the local graph with added self-connections; $ I_N $ is the identity matrix and $ \bold{\tilde{D}} $ is a diagonal matrix with $ \tilde{D}_{ii} = \sum_{j}\tilde{A}_{ij}$.
Our relational reasoning model is the stack of one Batch Normalization layer and four graph convolution layers activated by the ReLU function. We adopt softmax cross-entropy loss as the objective function for optimization. Similar to \cite{CVPR19_Linkage}, we only back-propagate the gradient for nodes on the 1-hop neighbors in training, because we just care about the linkage between a pivot and its 1-hop neighbors. For testing, we also only consider the classification on 1-hop nodes.

\subsection{Inference}
Given the text components, we group text components into text instances according to the reasoning results. We first apply thresholding to TR and TCR respectively, and then NMS is applied to reduce redundancy. To infer the likelihood of linkages between the pivot and its neighbors, we loop over all text components, constructing a local graph with each component as the pivot. Consequently, we obtain a set of edges weighted by the linkage likelihood. Finally, we use Breath First Search (BFS) to cluster and merge the linkages.

After we get the clustered text components, we sort the components for boundary generation. The text instance $ T $ can be represented as $ T  =\{D_0,...D_i,...,D_n\}$. The Min-Path algorithm is applied to search the shortest path through all text component centers, and then we sort $ T $  by searching results. For boundary generation, we just need to link the mid-point of the ordered top and bottom in ordered text components sequentially, as shown in Fig.$ \ $\ref{fig:fig2}.

\section{Experiments} \label{Experiments}
\subsection{Datasets}

\textbf{Total-Text}: It consists of $1,255$ training and $300$ testing complex images, including horizontal, multi-oriented, and curved text instances with polygon and word-level annotations.

\textbf{CTW-1500}: It consists of $1,000$ training and $500$ testing images. Every image has curved text instances, which are all annotated by polygons with 14 vertices.

\textbf{MSRA-TD500}: It consists of $500$ training and $200$ testing images, including English and Chinese scripts. This dataset is dedicated for detecting multi-lingual long texts of arbitrary orientations.

\textbf{ICDAR2015}: It consists of $1,000$ training images and $500$ testing images, including many multi-orientated and very small-scale text instances. The ground truth is annotated with word-level quadrangle.

\textbf{ICDAR2017}: It consists of $7,200$ training images, $1,800$ validation images and $9,000$ test images with texts in $9$ languages for multi-lingual scene text detection. The text instances are also annotated by quadrangle.

\subsection{Implementation Details}\label{trainstep}
The backbone of our network is the pre-trained VGG16 \cite{vgg} on ImageNet  \cite{ImageNet}. The training procedure mainly includes two steps:  pre-training our network on SynthText dataset with two epochs, and fine-tuning on specific benchmark dataset with $600$ epochs. In the pre-training stage, we randomly crop text regions, which will be resized to $512$. The batch size is set to $12$. Adam optimizer is applied to train our model with a learning rate $10^{-4}$.
In fine-tuning, for multi-scale training, we randomly crop the text region, and resize them to $640 \times 640$ (batch is $8$), $800 \times 800$ (batch is $4$), and $960 \times 960$ (batch is $4$), respectively. In fine-tuning, SGD optimizer is applied to train our model. The initial learning rate is $0.01$ and multiplied by $0.8$ after each $100$ epochs.
Also, the basic data augmentation techniques like rotations, crops, color variations, and partial flipping are applied. The hyper-parameters related to local graph are fixed during training and testing. Experiments are performed on single GPU (RTX-2080Ti), and PyTorch 1.2.0.

\begin{table}[htbp]
	\vspace{-0.0em}
	\begin{center}
	\renewcommand{\arraystretch}{1.0}
	\begin{tabular}{c||l|ccc}
		\hline
		\textbf{Datasets}& \textbf{Methods} &\textbf{R} &\textbf{P} &\textbf{H}\\
		\hline
		\multirow{2}*{Total-Text}&{baseline} &80.06 &85.45  & 82.67\\
		\cline{2-2}
		&baseline+\textbf{gcn} &83.11 &85.94 &84.50\\
		\hline
		\multirow{2}*{CTW1500}&{baseline} &80.57 &83.06&81.80\\
		\cline{2-2}
		&baseline+ \textbf{gcn} &81.45 &83.75&82.58\\
		\hline
		\multirow{2}*{TD500}&{baseline} &78.52 &83.24&80.81\\
		\cline{2-2}
		&baseline+ \textbf{gcn} &82.30 &88.05&85.08\\
		\hline

	\end{tabular}
    \end{center}%
	\vspace{-0.0em}
    \caption{Ablation study for relational reasoning network. “R”, “P” and “H” represent recall, precision and Hmean, respectively. For “baseline”, we adopt the intersection of TR and TCL, instead of the relationship learned by GCN, to group text patches. }
    \label{table:Ablation}
    \vspace{-0.5em}
\end{table}
\begin{table*}[htbp]
	\vspace{-0.0em}
	\begin{center}
	\renewcommand{\arraystretch}{1.0}
	\begin{tabular}{|c||c|c|c||c|c|c||c|c|c|}
		\hline
		\multicolumn{1}{|c||}{ \multirow{2}*{\textbf{Methods}} }& \multicolumn{3}{c||}{\textbf{Total-Text}} & \multicolumn{3}{c||}{\textbf{CTW-1500}} & \multicolumn{3}{c|}{\textbf{MSRA-TD500}}\\
		\cline{2-10}
		&\textbf{Recall}&\textbf{Precision}&\textbf{Hmean}&\textbf{Recall}&\textbf{Precision}&\textbf{Hmean}&\textbf{Recall}&\textbf{Precision}&\textbf{Hmean}\\
		\hline
		SegLink \cite{SegLink} &- &- &-&- &- &- &70.0 &86.0 &77.0\\
		MCN \cite{MCN} &- &- &- &- &- &- &79 &88 &83\\
		TextSnake \cite{TextSnake} &74.5 &82.7 & 78.4 &\textbf{85.3} &67.9 & 75.6 &73.9 &83.2 &78.3\\
		LSE$ ^{\dagger} $ \cite{CVPR19_LSA}&- &- &- &77.8 &82.7 & 80.1 &81.7& 84.2 &82.9\\
		ATTR$ ^{\dagger} $ \cite{CVPR19_ATRR}  &76.2 &80.9 & 78.5 &- &- &- &82.1 &85.2 & 83.6\\
		MSR$ ^{\dagger} $ \cite{MSR} &73.0 &85.2 & 78.6 &79.0 &84.1 & 81.5&76.7 &87.4 &81.7\\
		CSE \cite{CVPR19_CSE}&79.7&81.4&80.2&76.1&78.7&77.4&- &- &-\\
		TextDragon \cite{TextDragon}&75.7&85.6&80.3&82.8&84.5&83.6&- &- &-\\
		TextField \cite{TextField}&79.9&81.2&80.6&79.8&83.0&81.4&75.9&87.4&81.3\\
		PSENet-1s$ ^{\dagger} $ \cite{CVPR19_PSENet}&77.96&84.02&80.87&79.7&84.8& 82.2&- &- &-\\
		ICG \cite{SegLink++}&80.9&82.1& 81.5&79.8&82.8&81.3&- &- &-\\
		LOMO*$ ^{ \dagger }$ \cite{CVPR19_LOMO}&79.3&87.6&83.3&76.5&85.7&80.8&- &- &- \\
		CRAFT \cite{CRAFT}&79.9&87.6&83.6&81.1&86.0&83.5&78.2&\textbf{88.2}&82.9\\
		PAN $ ^{ \dagger } $ \cite{PSENet_v2}&81.0 &\textbf{89.3}  &85.0 &81.2 &\textbf{86.4} &83.7 &\textbf{83.8} &84.4 &84.1\\
		\hline
		\hline
		\textbf{Ours} & \textbf{84.93} &86.54  & \textbf{85.73}&83.02 &85.93& \textbf{84.45} &82.30 &88.05& \textbf{85.08}\\
		\hline
	\end{tabular}
    \end{center}%
	\vspace{-0.4em}
	 \caption{Experimental results on Total-Text, CTW-1500 and MSRA-TD500. The symbol $^*$  means the multi-scale test is performed. The symbol $ ^{\dagger}$ indicates the backbone network is not VGG16. The best score is highlighted in \textbf{bold}.}
	\label{table:tbSyn}
    \vspace{-1.2em}
\end{table*}
\begin{figure}
	\begin{center}
	\subfigure{
		\begin{minipage}[t]{0.48\linewidth}
			\centering
			\includegraphics[width=4.1cm,height=5cm]{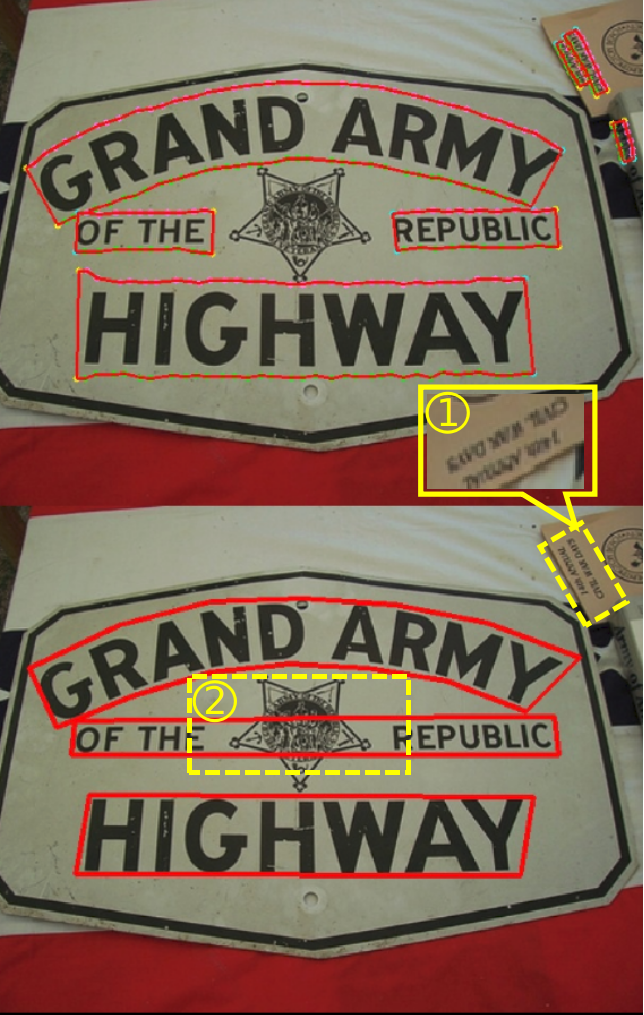}
		\end{minipage}
		\begin{minipage}[t]{0.5\linewidth}
			\centering
			\includegraphics[width=4.1cm,height=5cm]{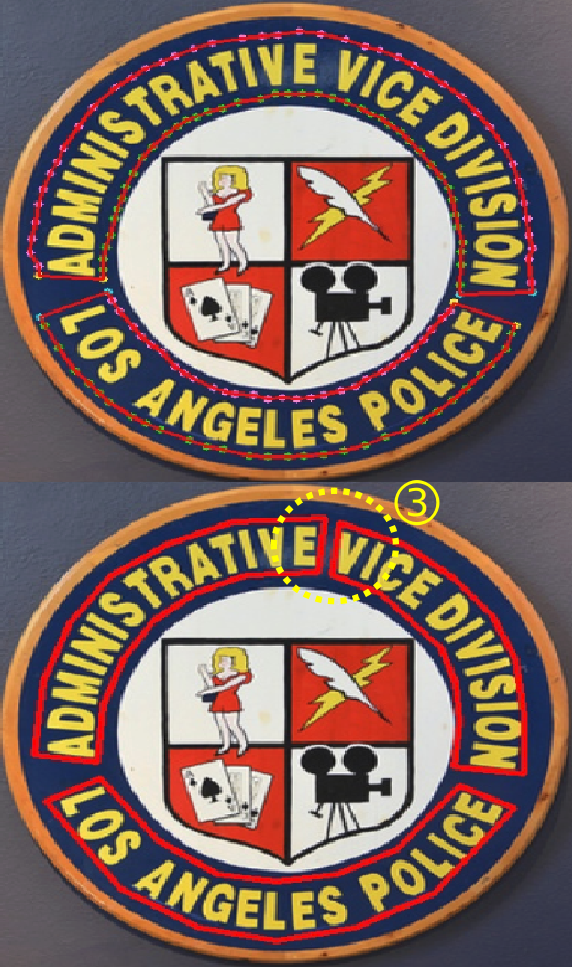}
		\end{minipage}
	}%
	\caption{ The representative samples with irregular labels on CTW-1500. Up row: the results of our method.  Bottom row: the ground truth of CTW-1500.}
	\label{fig:confuse}
    \end{center}
	\vspace{-1.5em}
\end{figure}
\subsection{Ablation Study}\label{exp_ablation_aam}
To verify the effectiveness of the relational reasoning network, We conduct ablation experiments on Total-Text, CTW1500 and MSRA-TD500. Tab.$ \, $\ref{table:Ablation} shows the experimental results on three datasets. For reducing the influence of data on the experimental results, we adopt the SynthText to pre-train model, and then we fine-tune it on Total-Text and CTW1500. Because MSRA-TD500 consists of English and Chinese, we use ICDAR2017-MLT to pre-train our network for MSRA-TD500. The longer sides of the images within Total-Text, CTW1500  and  MSRA-TD500 are restricted to $1,280$, $1,024$ and $640$, respectively, meanwhile keeping the aspect ratio. As shown in Tab.$ \, $\ref{table:Ablation}, the relational reasoning network achieves improvements by $1.83 \%$, $0.78 \%$ and  $4.27 \%$ in Hmean on Total-Text, CTW1500  and  MSRA-TD500, respectively. Remarkably, the recall of our method with relational reasoning network has improved significantly in all datasets ($3.05 \%$ on Total-Text, $ 0.88 \% $ on CTW1500, and $ 3.78 \% $ on MSRA-TD500). Our method coherently improves the detection performance on MSRA-TD500 abundant with long texts (recall $ 3.78\% $, precision $4.81 \%$, Hmean $4.27 \%$). The performance of our method on CTW1500 is not remarkable, because its annotations are sometimes confusing. The CTW1500 has no ”DO NOT CARE”, so some small texts and Non-English texts are not annotated, as shown in Fig.$ \, $\ref{fig:confuse} \textcircled{1}. Moreover, the text line annotations are confusing, as shown in Fig.$ \, $\ref{fig:confuse} \textcircled{2} and \textcircled{3}.

\subsection{Comparison with the state-of-the-arts}

\begin{figure*}[htbp]
\vspace{0.0em}
\centering
\subfigure[Total-Text]{
	\begin{minipage}[t]{0.25\linewidth}
		\centering
		\includegraphics[width=4.35cm,height=3cm]{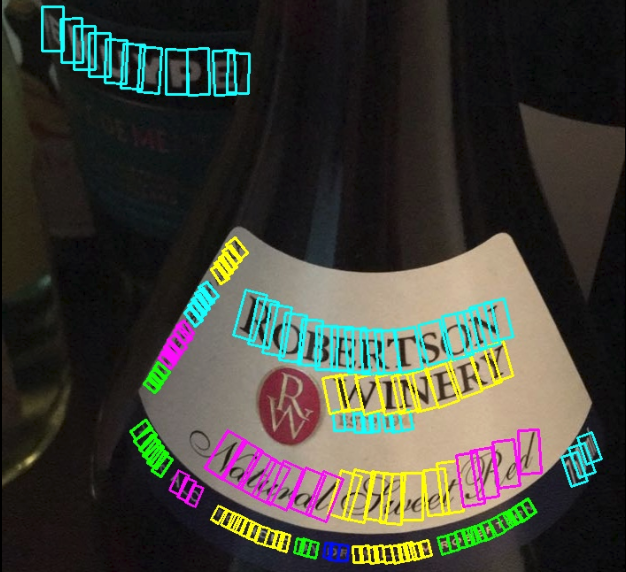}\\
		\includegraphics[width=4.35cm,height=3cm]{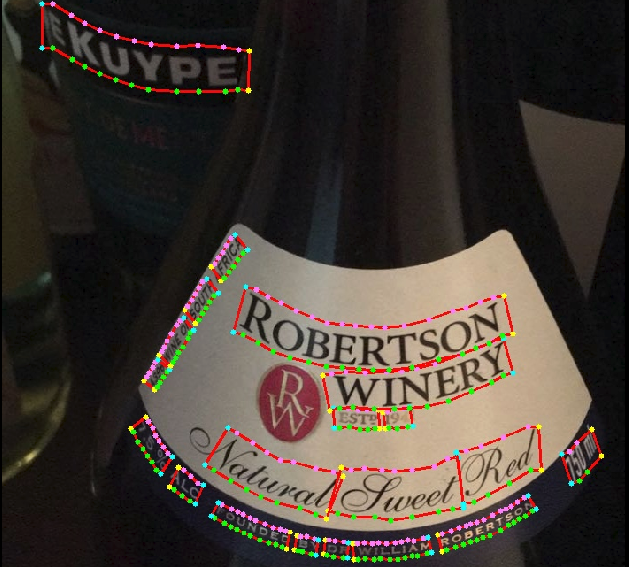}
	\end{minipage}%
}%
\subfigure[Total-Text]{
	\begin{minipage}[t]{0.25\linewidth}
		\centering
		\includegraphics[width=4.35cm,height=3cm]{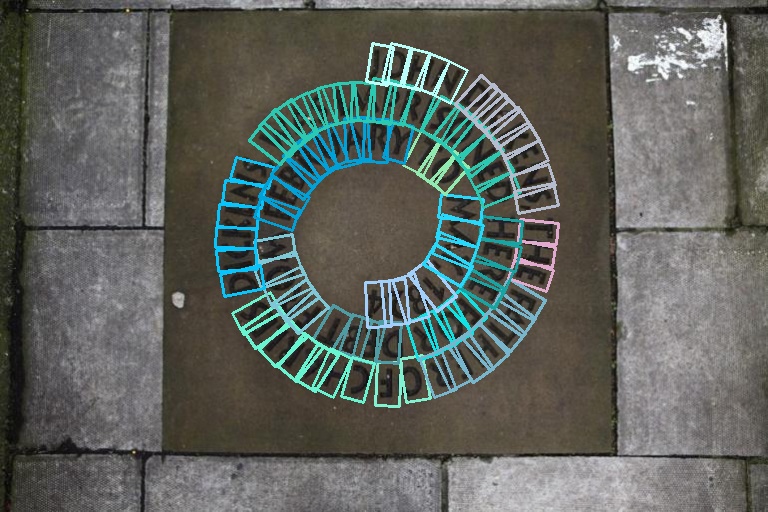}\\
		\includegraphics[width=4.35cm,height=3cm]{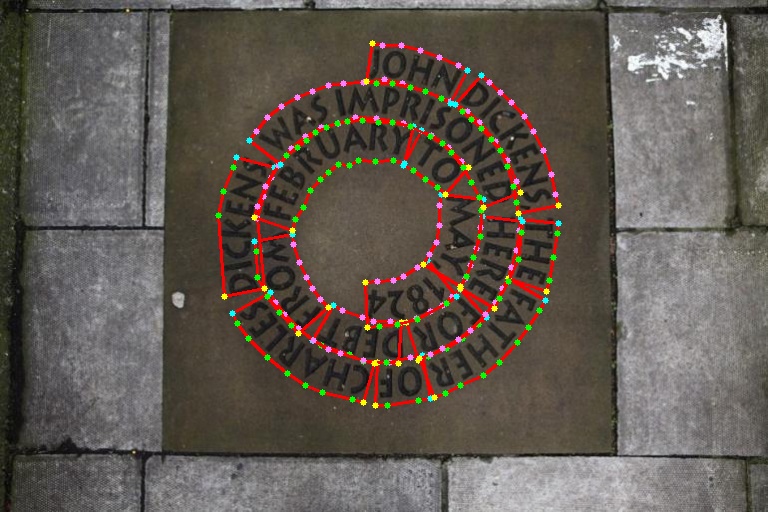}
	\end{minipage}%
}%
\subfigure[CTW1500]{
	\begin{minipage}[t]{0.25\linewidth}
		\centering
		\includegraphics[width=4.35cm,height=3cm]{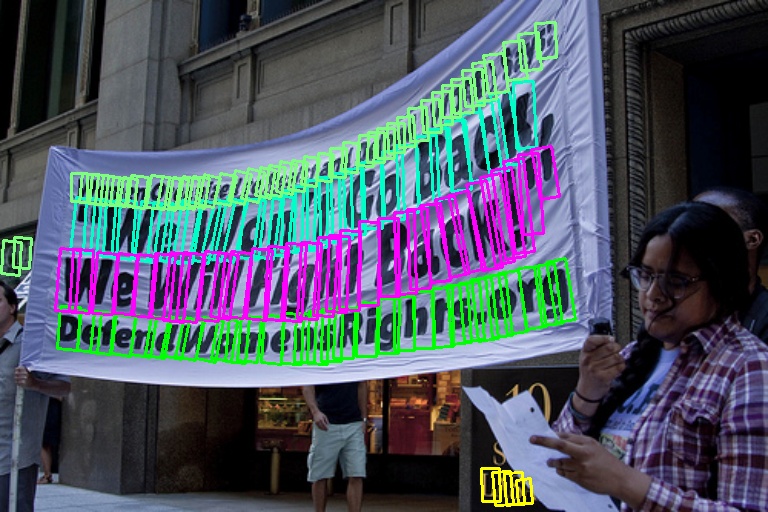}
		\includegraphics[width=4.35cm,height=3cm]{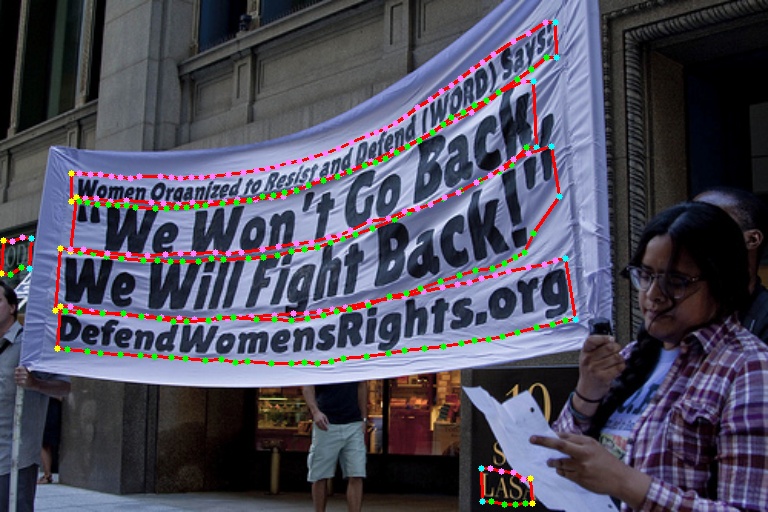}
	\end{minipage}%
}%
\subfigure[MSRA-TD500]{
	\begin{minipage}[t]{0.25\linewidth}
		\centering
		\includegraphics[width=4.35cm,height=3cm]{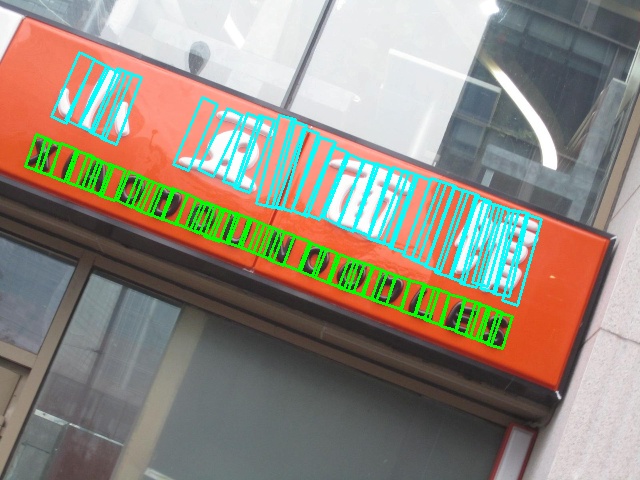}
		\includegraphics[width=4.35cm,height=3cm]{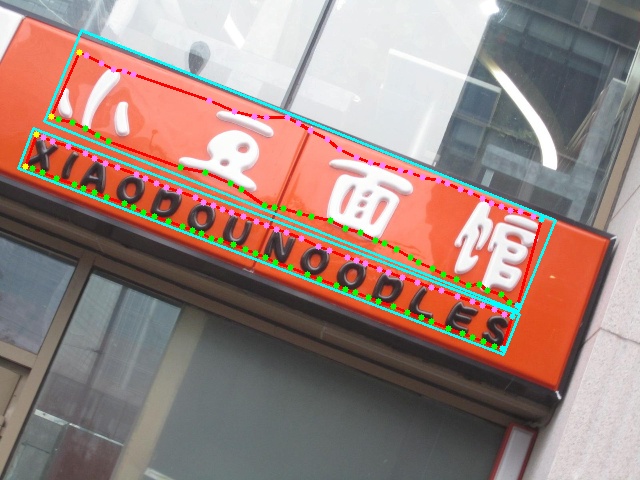}
	\end{minipage}%
}%
	
\centering
\caption{ Experimental results of our method. Up row: each column shows the results of GCN clustering on different datasets. Bottom row: each column shows the corresponding results of boundary generation.}
\label{fig:result_all}
\vspace{-1.2em}
\end{figure*}

\noindent\textbf{Polygon-Type Datasets.} Here, ICDAR2017-MLT is used to pre-train our model, and fine-tuning is only conducted on CTW1500 and Total-Text, separately. All experiments are performed with a single image resolution.

\textbf{Total-Text}. This dataset mainly contains curved and multi-oriented texts, annotated in word-level. In testing, we resize the shortest side to $512$ if it is less than $512$, and keep the longest side is not larger than $1,280$. Some visible results are listed in Fig.$ \, $\ref{fig:result_all} (a) (b). From Fig.$ \, $\ref{fig:result_all}, we can observe that our method precisely detects
word-level irregular texts, and it is can accurately
separate close text instances of arbitrary shapes. The quantitative results are shown in Tab.$ \ $\ref{table:tbSyn}. The proposed method achieves $85.73\%$ Hmean, significantly outperforming other methods.

\textbf{CTW1500}. This dataset mainly contains curved and multi-oriented texts, annotated in line-level. In testing, we resize the shortest side to $512$ if it is less than $512$, and keep the longest side is not larger than $1,024$. Some visible results are shown in Fig.$ \, $\ref{fig:result_all} (c) and Fig.$ \, $\ref{fig:confuse}. It indicates that the proposed method correctly detects the boundaries of arbitrary shape text precisely. The quantitative results are listed in Tab.$ \ $\ref{table:tbSyn}. Compared with the other stsate-of-the-art methods, our approach achieves promising in recall ($83.02\%$) and Hmean ($ 84.45\% $). Specifically, our method greatly outperforms TextSnake on CTW1500 and Total-Text, improves Hmean by $8.85\%$ and $6.6\%$ respectively.

\begin{table}[htbp]
    \begin{center}
	\renewcommand{\arraystretch}{1.1}
	\begin{tabular}{l||lll}
		\toprule
		\textbf{Methods}& \textbf{Recall} & \textbf{Precision} & \textbf{Hmean}\\
		\hline
		SegLink \cite{SegLink} &76.8 &73.1 &75.0\\
		MCN \cite{MCN} &72 &80 &76\\
		EAST$^*$ \cite{EAST} &78.3 &83.3 &80.7\\
		TextField \cite{TextField} &80.05 &84.3 &82.4\\
		TextSnake \cite{TextSnake} &84.9 &80.4 &82.6\\
		Textboxes++$^*$\cite{textboxes++} &78.5 &87.8 &82.9\\
		PixelLink \cite{PixelLink} &82.0 &85.5 &83.7\\
		FOTS$ ^{\dagger} $ \cite{FOTS}&82.04 &88.84 &85.31\\
		PSENet-1s$ ^{\dagger} $  \cite{CVPR19_PSENet} &84.5 &86.92 & 85.69\\
		LSE$ ^{\dagger} $  \cite{CVPR19_LSA} &\textbf{85.0} &88.3 &86.6 \\
		ATRR$ ^{\dagger} $  \cite{CVPR19_ATRR} &83.3 &\textbf{90.4} &86.8\\
		CRAFT \cite{CRAFT} &84.3 &89.8 &\textbf{86.9} \\
		\hline
		\textbf{Ours} &84.69& 88.53& 86.56\\
		\bottomrule
	\end{tabular}
    \end{center}%
	\vspace{-0.5em}
	\caption{Experimental results on ICDAR2015.}
	\label{table:ICDAR15}
    \vspace{-1.6em}
\end{table}

\vspace{0.5em}
\noindent\textbf{Quadrilateral-Type Datasets.} For fairly comparison, we adopt IC17 to pre-train our model, then fine-tune it on IC15 and TD500, separately. However, these datasets are evaluated with rectangular boxes, hence we need to convert the detection results into rectangular boxes. Therefore, we shrink the text instance by $0.05$, and take the smallest circumscribed rectangle for evaluation.

\textbf{MSRA-TD500}. This dataset contains lots of long texts and text scales vary significantly. In testing, we resize the shortest side to $512$ if it's less than $512$, and keep the longest side isn't larger than $640$. Fig.$ \, $\ref{fig:result_all} (d) are some representative results. The proposed method
successfully detects long text lines of arbitrary orientations
and sizes. The quantitative comparisons with other methods on this dataset is listed in Tab.$ \, $\ref{table:tbSyn}. Notably, our method achieves $85.08\%$ on Hmean, significantly outperforms other methods.

\textbf{ICDARs (IC15, IC17)}. Considering IC15 contains many low resolution and many small text instances. The instance balance \cite{PixelLink} is applied to assist training. The IC17 contains multilingual scene text and the annotations are given in word-level. In inference, we adjust the size of test images appropriately. For IC15, we resize the shortest side to $960$ if it is less than $960$, and keep the longest side is not larger than $1,960$. For IC17, we resize the shortest side to $512$ if it is less than $512$, and keep the longest side is not larger than $2,048$. The quantitative results are listed in Tab.$ \ $\ref{table:MLT} and Tab.$ \ $\ref{table:ICDAR15}.
Apparently, our method achieves $86.56\%$ Hmean on IC15 and $67.31\%$ Hmean on IC15. The proposed method achieves competitive results against the state-of-the-art methods.

\begin{table}[tbp]
	\begin{center}
	\renewcommand{\arraystretch}{1.1}
	\begin{tabular}{l||lll}
		\toprule
		\textbf{Methods}& \textbf{Recall}& \textbf{Precision} & \textbf{Hmean}\\
		\hline
		SARI FDU RRPN\cite{RRPN} &55.50 &71.17 &62.37\\ 
		He et al. \cite{MOML} &57.9 &76.7  &66.0\\ 
		Border$ ^{\dagger} $ \cite{ASTD} &60.6 &73.9  &66.6\\ 
		Lyu et al. \cite{corner} &55.6 &\textbf{83.8} &66.8\\
		FOTS$ ^{\dagger} $ \cite{FOTS} &57.51 &80.95 &67.25\\
		LOMO$ ^{\dagger} $ \cite{CVPR19_LOMO} &60.6 &78.8 &\textbf{68.5}\\
		\hline
		\textbf{Ours} &\textbf{61.04} &74.99 & 67.31\\
		\bottomrule
	\end{tabular}
	\end{center}%
	\vspace{-0.5em}
	\caption{Experimental results on ICDAR17 MLT.}
	\label{table:MLT}
    \vspace{-1.5em}
\end{table}


\section{Conclusion} \label{Conclusion}
In this paper, we propose a novel CC-based method for arbitrary shape scene text detection. The proposed method adopts a spectral-based graph convolution network learn linkage relationship between the text components, and use this information to guide post-processing to connect components to text instances correctly. Experiments on five benchmarks show that the proposed method not only has good performance for arbitrary shape text detection, but also good for oriented and multilingual text. In the future, we are interested in developing an end-to-end text reading system for text of arbitrary shapes with  graph network.

\vspace{1.0em}
\noindent\textbf{Acknowledgements.} This work was supported by National Key R\&D Program of China (No.2019YFB1405990), Beijing Natural Science Foundation (No.4194084), China Postdoctoral Science Foundation (No.2018M641199) and Fundamental Research Funds for the Central Universities (No. FRF-TP-18-060A1).

{\small
	\bibliographystyle{ieee_fullname}
	\bibliography{main}
}

\end{document}